\documentclass[11pt]{article}
\pdfoutput=1
\usepackage{microtype}
\usepackage{graphicx}
\usepackage{subfigure}
\usepackage{booktabs}
\usepackage{amssymb}
\usepackage{bm}
\usepackage{bbm}
\usepackage{amsmath}
\usepackage{amsthm}
\usepackage{xcolor}
\usepackage{textcomp}
\usepackage{multirow}
\usepackage{caption}
\usepackage{mathtools}
\usepackage[margin=1in]{geometry}
\usepackage{authblk}
\usepackage[numbers]{natbib}
\usepackage{xspace}
\usepackage{comment}
\usepackage{makecell}
\usepackage{thmtools, thm-restate}
\usepackage{verbatim}
\usepackage{todonotes}
\newcolumntype{L}[1]{>{\raggedright\arraybackslash}p{#1}}
\usepackage{tikz}
\usetikzlibrary{arrows}
\usetikzlibrary{positioning}
\usepackage{tcolorbox}
\usepackage{makecell} % Add this in your preamble

\tikzset{
  treenode/.style = {align=center, inner sep=0pt, text centered,
    font=\sffamily},
  arn_n/.style = {treenode, circle, black, font=\sffamily\bfseries, draw=black,
    fill=white, text width=1.5em},
  arn_r/.style = {treenode, circle, black, font=\sffamily\bfseries, draw=black,
    fill=white, text width=1.0em},
  arn_x/.style = {treenode, rectangle, draw=black,
    minimum width=0.5em, minimum height=0.5em}
}

\usepackage{hyperref}

\newcommand{\oThreeScore}{71.7}
\newcommand{\codemonkeysCoverage}{69.8}
\newcommand{\codemonkeysScore}{57.4}
\newcommand{\codemonkeysRandomScore}{45.8}
\newcommand{\codemonkeysCost}{2300}
\newcommand{\barrelCoverage}{80.8}
\newcommand{\barrelScore}{66.2}
\newcommand{\barrelRandomScore}{60.9}
\newcommand{\bestScoreInEnsemble}{62.8}

\usepackage{bm}
\usepackage{enumitem}

\newcommand*{\ShowNotes}{}

\ifdefined\ShowNotes
  \newcommand{\colornote}[3]{{\color{#1}\bf{#2 #3}\normalfont}}
\else
  \newcommand{\colornote}[3]{}
\fi

\definecolor{darkred}{rgb}{0.7,0.1,0.1}
\definecolor{darkgreen}{rgb}{0.1,0.5,0.1}
\definecolor{cyan}{rgb}{0.7,0.0,0.7}
\definecolor{dblue}{rgb}{0.2,0.2,0.8}
\definecolor{maroon}{rgb}{0.76,.13,.28}
\definecolor{burntorange}{rgb}{0.81,.33,0}
\definecolor{royalpurple}{rgb}{0.47,.31,0.66}

\ifdefined\ShowNotes
  
\else
  
\fi

\newif\ifarxiv

\title{CodeMonkeys: Scaling Test-Time Compute for Software Engineering}

\author[$\dagger$]{Ryan Ehrlich$^*$}
\author[$\dagger\ddagger$]{Bradley Brown$^*$}
\author[$\dagger$]{Jordan Juravsky$^*$}
\author[$\ddagger$]{Ronald Clark}
\author[$\dagger$]{Christopher R{\'e}}
\author[$\dagger$]{Azalia Mirhoseini}  \affil[$\dagger$]{Department of Computer Science, Stanford University}
\affil[$\ddagger$]{University of Oxford}

\affil[ ]{\normalsize\texttt{rehrlich@stanford.edu, bradley.brown@cs.ox.ac.uk, jbj@stanford.edu, ronald.clark@cs.ox.ac.uk, chrismre@stanford.edu, azalia@stanford.edu}\vspace{-0.2cm}}

\date{January 24, 2025}

\begin{document}

\maketitle

\begin{abstract}
Scaling test-time compute is a promising axis for improving LLM capabilities.
However, test-time compute can be scaled in a variety of ways, and effectively combining different approaches remains an active area of research.
Here, we explore this problem in the context of solving real-world GitHub issues from the SWE-bench dataset. 
Our system, named CodeMonkeys,  allows models to iteratively edit a codebase by jointly generating and running a testing script alongside their draft edit. 
We sample many of these multi-turn trajectories for every issue to generate a collection of candidate edits. 
This approach lets us scale ``serial'' test-time compute by increasing the number of iterations per trajectory and ``parallel'' test-time compute by increasing the number of trajectories per problem. 
With parallel scaling, we can amortize up-front costs across multiple downstream samples, allowing us to identify relevant codebase context using the simple method of letting an LLM read every file.
In order to select between candidate edits, we combine voting using model-generated tests with a final multi-turn trajectory dedicated to selection.
Overall, CodeMonkeys resolves \codemonkeysScore\% of issues from SWE-bench Verified using a budget of approximately \codemonkeysCost\ USD.
Our selection method can also be used to combine candidates from different sources. Selecting over an ensemble of edits from existing top SWE-bench Verified submissions obtains a score of \barrelScore\% and outperforms the best member of the ensemble on its own.
We fully release our code and data at \url{https://scalingintelligence.stanford.edu/pubs/codemonkeys/}.

{\let\thefootnote\relax\footnotetext[0]{{$^*$ Equal Contribution. Work done by BB as a visiting researcher at Stanford.}}}

\end{abstract}

\section{Introduction}

\begin{figure*}[t]
    \centering
    \includegraphics[width=\textwidth]{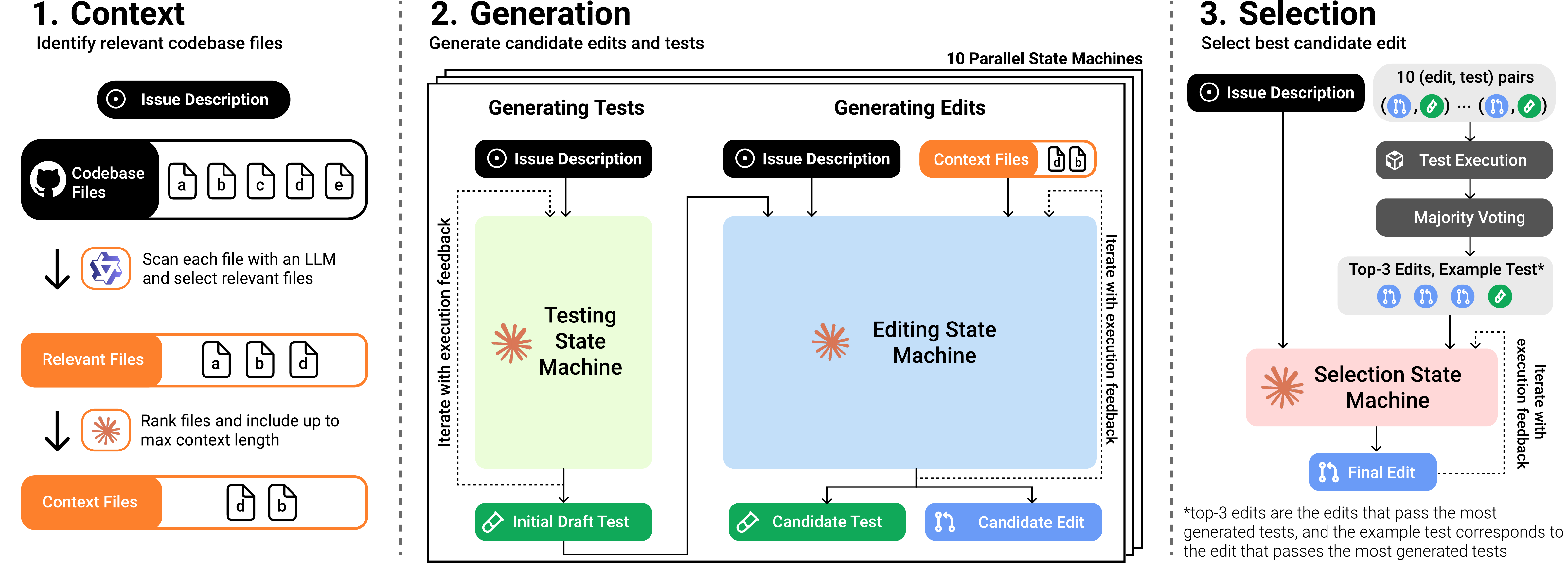}
    \caption{Overview of the CodeMonkeys system. \textbf{Left:} We retrieve codebase context by using models to first identify relevant files and then rank them relative to each other. \textbf{Middle:} We generate a codebase edit and testing script using a pair of multi-turn state machines that iterate based on execution feedback. We run these state machines multiple times in parallel to generate 10 edits and tests for every issue. \textbf{Right:} We select between candidate edits by identifying the candidates that pass the most generated tests and asking a model to decide between these top candidates. For details about our system's three state machines, see Figure~\ref{fig:sm_detail}.}
    \label{fig:banner}
    \vspace{-0.5cm}
\end{figure*}

\begin{table}[htbp]
\centering
\begin{tabular}{lccccccr}
\toprule
 &
 \multicolumn{4}{c}{\textbf{Claude Sonnet-3.5 API Costs}} &
 \multicolumn{1}{c}{\textbf{Local Costs}} &
 \multicolumn{1}{c}{\textbf{Total Cost}} 
 \\
\cmidrule(lr){6-6} \cmidrule(lr){2-5} \cmidrule(lr){7-7}
\textbf{Stage}  &
\textbf{Input} &
\textbf{Output} &
\multicolumn{2}{c}{\textbf{Input Cache}} &
\textbf{Qwen-2.5} &
\textbf{USD (\%)} \\
\cmidrule(lr){4-5}
&
&
&
\textbf{Read} &
\textbf{Write} &
&  \\
\toprule
Relevance  & 0.00  & 0.00 & 0.00 & 0.00 & 334.02  & 334.02 (14.6\%) & \\
Ranking  & 0.00  & 11.92 & 1.10 & 6.90 & 0.00  & 19.92 (0.9\%) & \\
\midrule
Gen. tests  & 10.60  & 295.15 & 21.60 & 112.64 & 0.00  & 439.99 (19.2\%) & \\
Gen. edits  & 14.67  & 353.95 & 636.82 & 360.58 & 0.00  & 1366.02 (59.6\%) & \\
\midrule
Selection  & 0.52  & 51.12 & 15.17 & 65.14 & 0.00  & 131.95 (5.8\%) & \\
\midrule
\textbf{Total}  & 25.79  & 712.14 & 674.69 & 545.26 & 334.02  & 2291.90 (100.0\%) & \\
\bottomrule
\end{tabular}

\caption{Breaking down the costs of running CodeMonkeys on all GitHub issues from SWE-bench Verified. All costs are in USD. Our system uses two LLMs: a primary model used for the ranking, generation, and selection stages (we use the Claude 3.5 Sonnet API \cite{claude35sonnet}), and a cheaper model used for scanning codebases to identify relevant files (we run Qwen2.5-Coder-32B-Instruct \cite{hui2024qwen25coder} locally). For measuring costs with the Claude API, we use prices of \$3/million input tokens, \$0.3/million cache read tokens, \$3.75/million cache write tokens, and \$15/million output tokens. For details about estimating local inference costs, see Appendix \ref{sec:local_costs}.}

\label{tab:llm-costs}
\end{table}

The abilities of large language models (LLMs) to solve increasingly complex coding tasks have rapidly improved \cite{chen2021evaluatinglargelanguagemodels, li2023starcodersourceyou, claude35sonnet}. Modern LLMs can outperform some human contestants in programming competitions \cite{Li_2022_alphacode, alphacode2} and have become increasingly popular programming assistants \cite{ray2023microsoftcopilot, anysphere2025seriesbcursor}. 
LLMs are also improving at software engineering tasks such as solving real-world GitHub issues, as measured by the SWE-bench dataset \cite{jimenez2024swebenchlanguagemodelsresolve}. 
A major driver of progress has come from increasing the amount of compute and data used for model training, which has reliably led to improvements in model capabilities~\cite{hestness2017deeplearningscalingpredictable, kaplan2020scaling, hoffmann2022trainingcomputeoptimallargelanguage}. However, the costs of continuing to scale model training are becoming prohibitively expensive for most organizations \cite{grattafiori2024llama3herdmodels}.

An alternative avenue for further improving model capabilities, including for coding tasks such as SWE-bench, is to scale test-time compute \cite{Li_2022_alphacode, monkeys, snell2024scalingllmtesttimecompute, wu2024inferencescalinglawsempirical}. This type of scaling increases the amount of computation expended during inference in order to produce higher-quality solutions.
One approach to scaling test-time compute involves letting models deliberate for longer before outputting a final answer. This ``serial'' scaling can take the form of a chain-of-thought \cite{nye2021workscratchpadsintermediatecomputation, wei2023chainofthoughtpromptingelicitsreasoning} where models use many tokens of compute to reason through a problem, or through multi-turn interaction, where models iteratively respond to external feedback such as code execution results~\cite{yao2023reactsynergizingreasoningacting,
wang2024openhandsopenplatformai, xia2024agentlessdemystifyingllmbasedsoftware, ruan2024specrovercodeintentextraction}.
Alternatively, test-time compute can be scaled in ``parallel'' by sampling multiple candidate solutions to a problem \cite{wang2023selfconsistency, Li_2022_alphacode, lightman2023letsverifystepstep, flash2}.
In our previous work, Large Language Monkeys \cite{monkeys}, we found that coverage -- the fraction of problems in a dataset that are solved using any sample that was generated -- often increases log-linearly with the number of samples, including when solving issues from SWE-bench.

While these coverage results provide encouraging evidence that scaling parallel test-time compute may be beneficial for SWE-bench, they do not directly provide an actionable recipe for using it to improve issue solve rates. In particular, generating multiple candidates introduces the problem of needing to select a final answer among them. Benefiting from high coverage requires a selection method that can distinguish between correct and incorrect answers. Additionally, in Large Language Monkeys, we used an off-the-shelf SWE-bench framework (Moatless Tools \cite{moatless}) designed for generating a single solution. We generated multiple candidate edits by simply sampling from this framework repeatedly with a positive temperature.
This raises the question: how would one design a system differently if benefiting from test-time compute scaling was a primary consideration?

The core contribution of this work is to explore this idea, presenting a system for solving SWE-bench instances designed specifically around scaling test-time compute (Figure~\ref{fig:banner}).
We segment the resolution of an issue into three major steps: 1) identifying relevant codebase context, 2) generating candidate codebase edits for resolving the issue, and 3) selecting among these candidate edits \cite{xia2024agentlessdemystifyingllmbasedsoftware}.
We scale serial compute when generating a codebase edit by enforcing that models also write a testing script alongside their edit, allowing them to iteratively revise their edits and tests in response to execution feedback. We scale parallel test-time compute by sampling many of these (edit, test) pairs for every SWE-bench issue.
This combination of scaling achieves \codemonkeysCoverage\% coverage on SWE-bench Verified.
Interestingly, we find that different ways of allocating an inference budget between serial and parallel scaling often lead to similar coverage values.
Additionally, our use of parallel scaling lets us amortize the cost of identifying relevant codebase context across multiple downstream samples.
We adopt the simple method of letting an LLM scan every file, which contributes only 15\% to total costs when run up-front once per issue.

To select between candidate codebase edits, we explore methods based on voting with model-generated tests \cite{ruan2024specrovercodeintentextraction, xia2024agentlessdemystifyingllmbasedsoftware} and directly using models to do selection \cite{ruan2024specrovercodeintentextraction, flash2}.
We find that a combination of these two approaches works best, where test-based voting filters down the initial pool of candidates and a model selects among the remaining edits.
Moreover, we find that model-based selection can be further improved with more serial compute by letting models write and run tests to distinguish between candidates.
With this selection method, CodeMonkeys achieves an overall score of 57.4\% on SWE-bench Verified (Table~\ref{tab:scores}) while spending approximately 2300 USD on LLM inference (Table~\ref{tab:llm-costs}). 

We also show that our approach to selection can be used to effectively combine generations from heterogeneous sources. We demonstrate this by assembling the ``Barrel of Monkeys'': an expanded pool of candidate edits that include the top-4 submissions on the SWE-bench Verified leaderboard. Selecting over this ensemble, which has a coverage of \barrelCoverage\%, yields a score of \barrelScore\% - higher than the top-performing ensemble submission of \bestScoreInEnsemble\% and only 5.5\% below the reported score of o3 (\oThreeScore\%).
We release our code along with all generated samples at \url{https://scalingintelligence.stanford.edu/pubs/codemonkeys/}.

\section{Designing a SWE-bench Solver that Scales Test-Time Compute}
\label{sec:method}

\begin{figure*}[t]
    \centering
    \includegraphics[width=\textwidth]{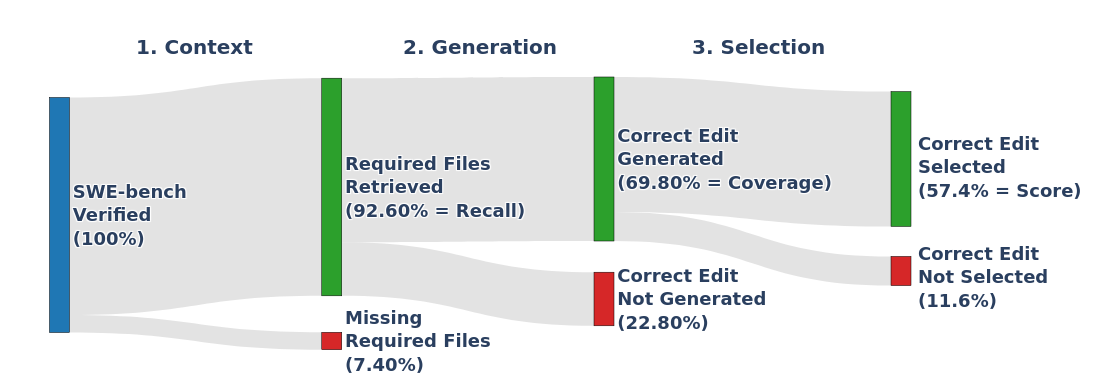}

    \caption{Measuring CodeMonkeys performance across the three subtasks we identify in Section~\ref{sec:method} (context, generation, and selection). Note that modifying the approach to one subtask can influence the performance on other subtasks as well. For example, generating more candidate edits could increase coverage but make selection harder.}
    \label{fig:problem_resolution_flow}
    \vspace{-0.5cm}
\end{figure*}

Each SWE-bench instance consists of an issue description and a corresponding code repository. The objective is to edit one or more files in the codebase in order to resolve the issue. Edits can be automatically scored for correctness using the repository's testing suite. In our work, we focus on the Verified \cite{chowdhury2024swebenchverified} split of SWE-bench, which contains instances that human annotators have classified as ``solvable'' (i.e. by having unambiguous issue descriptions and test suites that do not filter out correct solutions).
We decompose solving an instance from SWE-bench Verified into three sequential subtasks (Figure~\ref{fig:banner}):

\begin{enumerate}
    \item \textbf{Context:} Can we identify the codebase files that need to be edited\footnote{When calculating recall, we determine if a file needs to be edited by checking if the file is edited in the official issue solution provided in the SWE-bench dataset. This calculation does not account for the existence of alternative solutions that resolve the issue by editing a different set of files.} and put them in the context window? We can measure outcome of this subtask with \textbf{recall:} the fraction of problems where all needed files have been identified.
    \item \textbf{Generation:} Can we produce a correct codebase edit among any of our sampled candidates? We can measure this outcome of this subtask with \textbf{coverage:} the fraction of problems where at least one generated edit is correct.
    \item \textbf{Selection:} Can we choose a correct codebase edit from our collection of candidates? After completing this subtask, we can measure our final \textbf{score:} the fraction of problems in the dataset that are resolved by the edit our system submits.
\end{enumerate}

We describe our approach to each subtask below, with per-subtask metrics reported in Figure~\ref{fig:problem_resolution_flow} and a cost breakdown of our system in Table~\ref{tab:llm-costs}. Unless otherwise noted, all parts of our system use \verb|claude-3-5-sonnet-20241022|~\cite{claude35sonnet}. We present additional results using DeepSeek-V3~\cite{deepseekai2024deepseekv3technicalreport} in Appendix~\ref{app:deepseek}.

\subsection{Identifying Relevant Codebase Context}
\label{sec:context}

\begin{figure*}[t]
    \centering
    \includegraphics[width=\textwidth]{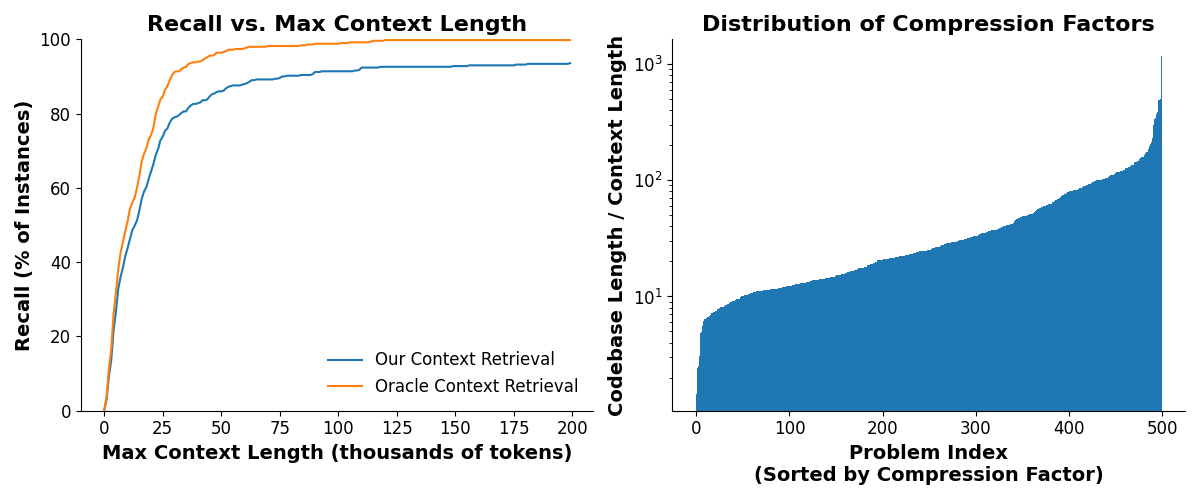}
    \caption{\textbf{Left}: Measuring recall (the fraction of SWE-bench problems with context windows that contain all needed files) as we increase the context window size limit.
    With the 128k token limit that we use for later experiments, 92.6\% of instances have the correct files in context. \textbf{Right}: Visualizing the distribution of context compression factors across SWE-bench problems, i.e. the ratio between the cumulative token count of files scanned by the relevance model and the cumulative token count of files we include after relevance + ranking.}
    \label{fig:context_recall}
    \vspace{-0.5cm}
\end{figure*}

\begin{figure*}[t]
    \centering
    \includegraphics[width=\textwidth]{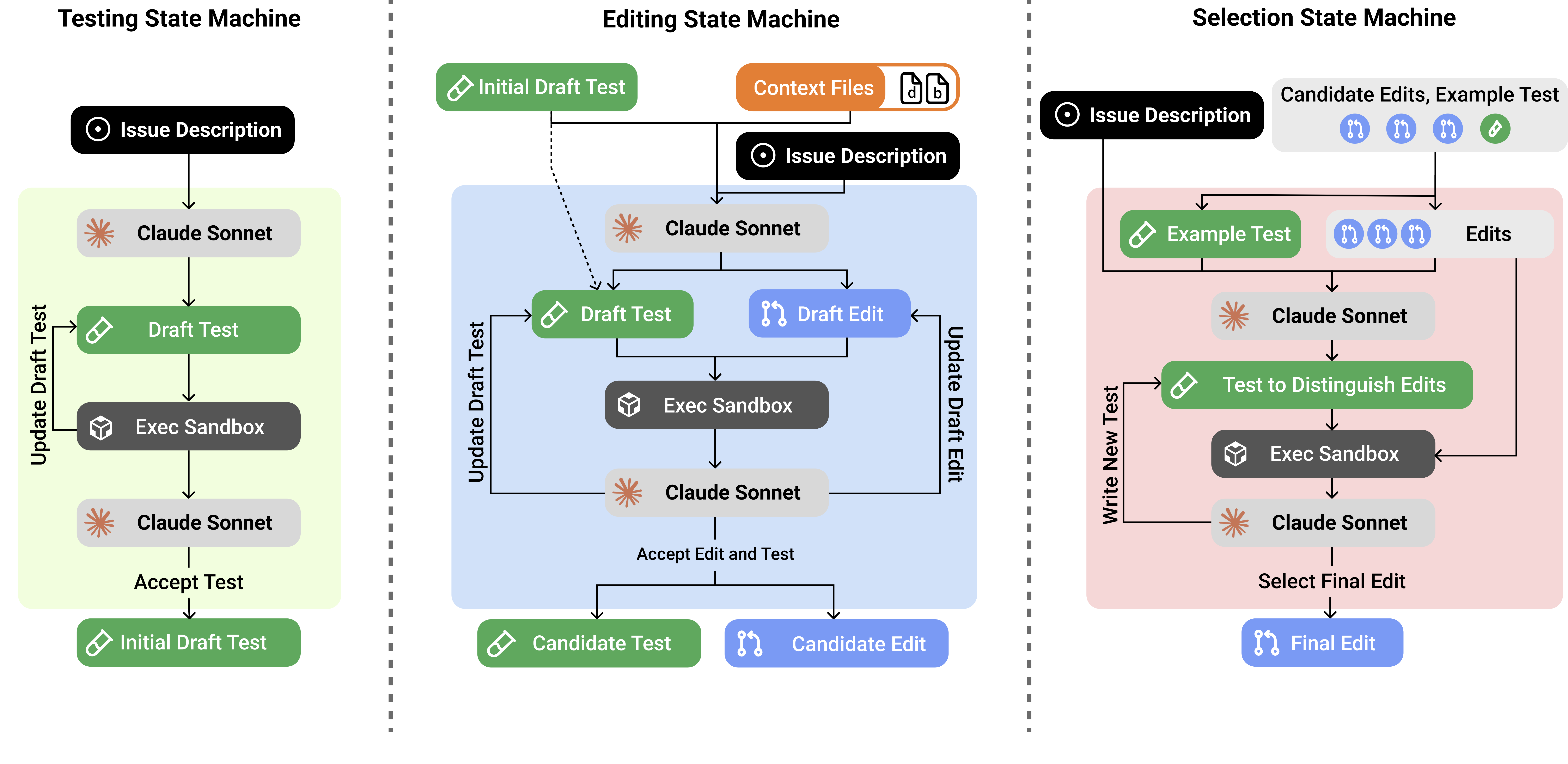}
    \caption{Details of the CodeMonkeys state machines. The \textbf{Testing State Machine} iteratively generates an initial draft of a testing script based on execution feedback from running the test on the codebase before any edits are applied. The \textbf{Editing State Machine} first generates an initial edit conditioned on the codebase context and the output test of the Testing State Machine. Then, it refines both the test and edit draft based on execution feedback from running the test before and after the edit is applied. The \textbf{Selection State Machine} first generates a test to distinguish between the top 3 candidate edits that pass the most testing scripts. Then, based on execution feedback of running this test with all of the candidate edits and on the codebase without edits, chooses to either create a new test script to further differentiate between the edits or selects a final edit.}
    
    \label{fig:sm_detail}
    \vspace{-0.5cm}
\end{figure*}

One of the key challenges when solving SWE-bench instances is managing the large volume of input context. Most SWE-bench codebases contain millions of tokens worth of context.
This exceeds the context lengths of most available models and would moreover be prohibitively expensive to process using frontier models. Existing approaches to managing SWE-bench context include using embedding models \cite{moatless, xia2024agentlessdemystifyingllmbasedsoftware}, iterative expansion from a file tree \cite{xia2024agentlessdemystifyingllmbasedsoftware}, and giving models tools for browsing files \cite{yang2024sweagentagentcomputerinterfacesenable, schluntz2024raising}. 

With our system, we know that we will generate a collection of candidate edits for every instance. Therefore, by choosing to share codebase context across all downstream edits, we can amortize the cost of context generation. 
This observation enables the simple approach of letting a model (we use Qwen2.5-Coder-32B-Instruct \cite{hui2024qwen25coder}) read every file in the codebase\footnote{We only include Python files in our scan and exclude files inside of testing directories.}, decide whether each is relevant to the target issue, and only include relevant files in our context window~\cite{arora2023languagemodelsenablesimple}.
Performing this codebase-wide scan once per instance contributes less than 15\% to our total system costs (Table~\ref{tab:llm-costs}) and on average processes 2.94M tokens per problem. If we did not amortize this scan and instead reran it for each of the 10 edits we generate per problem, it would become the most expensive step.
Sharing context across multiple downstream edits additionally saves costs by increasing hit rates when using prompt caching.

Even after the relevance filter, many problems still have contexts that are too long. To compress context further, we perform a model-based ranking procedure to order files by importance. First, as part of the initial codebase scan, we generate a concise summary for every file flagged as relevant that describes how the file relates to the target issue. We then construct a ranking prompt that includes each relevant file's name, summary, and token count.
We ask the ranking model to include approximately 60,000 tokens of context in its ranking.
Since the Claude API can be non-deterministic (even at temperature 0), we generate three completions from the ranking prompt and construct a final ranking by considering each file's average rank across the three repetitions. We construct a context window from this combined ranking by including the full contents all ranked files up to a limit of 128,000 tokens. On average, this leads to 74,570 tokens of codebase context, corresponding to an average 50.5x reduction in context size when compared with including every file that was assessed for relevance.

\subsection{Generating Candidate Codebase Edits with Corresponding Tests}
\label{sec:generation}

\begin{figure*}[t]
    \centering
    \includegraphics[width=\textwidth]{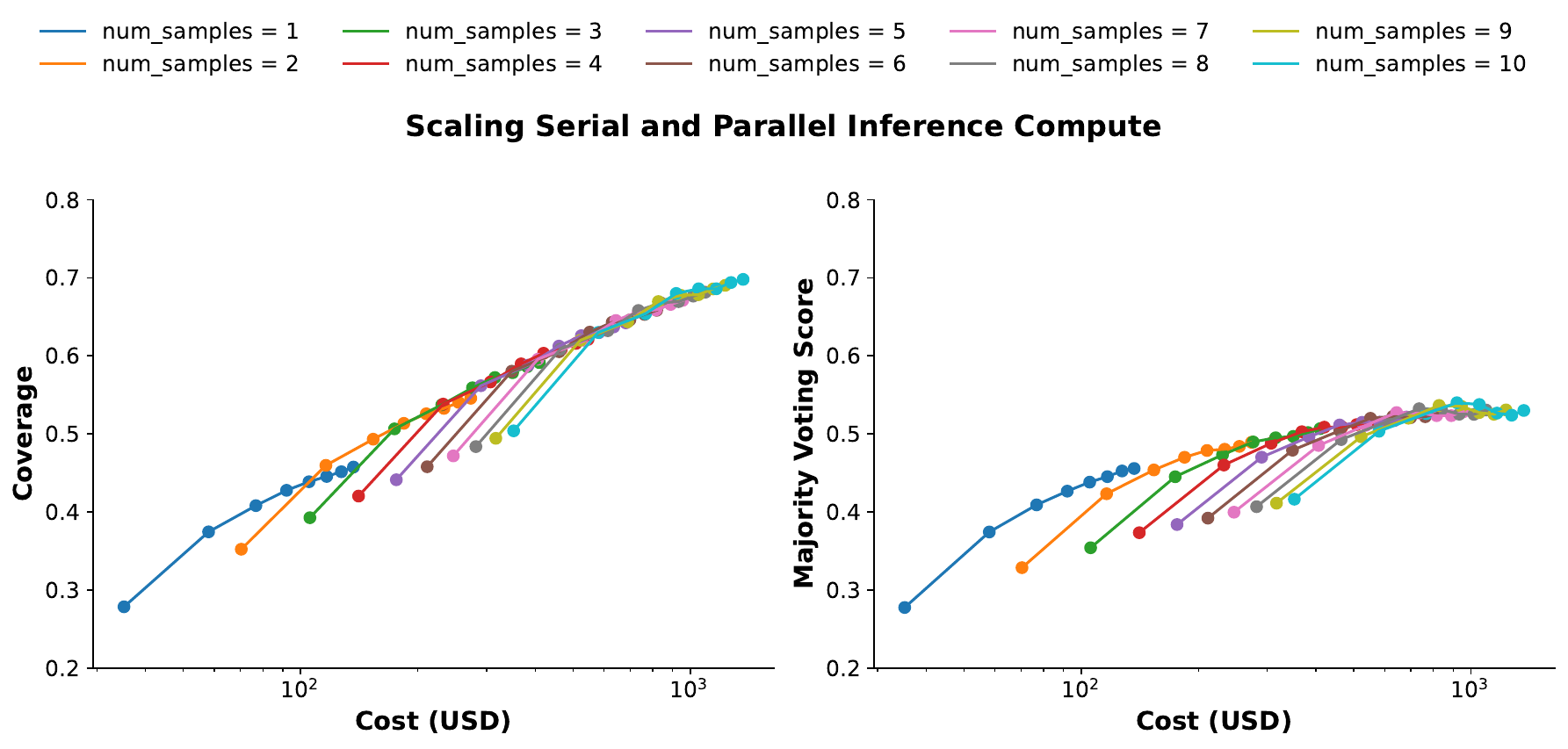}
    \caption{Measuring coverage (left) and score when using majority-voting selection (right) as we sweep over the number of serial iterations per editing state machine and the number of parallel state machines sampled per problem. Each colored curve corresponds to a different number of parallel state machines, and the dots along each curve correspond to increased numbers of sequential iterations per state machine. The first few serial iterations have a large impact on improving performance. However, past that point, different configurations with similar costs lead to similar performance, particularly for coverage.}
    \label{fig:coverage_scaling}
    \vspace{-0.5cm}
\end{figure*}

With relevant parts of the codebase identified, we can begin generating candidate codebase edits for solving the target issue.
We adopt a state machine abstraction~\cite{moatless} to model a multi-turn exchange where a model iteratively edits the target codebase in response to execution feedback.
Additionally, like in Large Language Monkeys~\cite{monkeys}, we run multiple independent state machines for every SWE-bench instance, using a positive sampling temperature to introduce diversity across candidates. This approach provides us with two straightforward ways to scale test-time compute:

\begin{enumerate}
    \item We can scale serial compute by increasing the maximum number of iterations performed per state machine\footnote{Precisely, we limit the number of model completions per state machine. The model's initial generation and corrections to previous malformatted responses count against this limit.}.
    \item We can scale parallel compute by increasing the number of independent state machines run per instance.
\end{enumerate}

Importantly, we require that models generate and revise a test jointly with each edit. These tests are structured as standalone Python scripts that attempt to reproduce the GitHub issue and communicate their results using exit codes. Forcing models to write executable scripts alongside edits provides models with richer feedback to guide their iteration and scale serial compute more effectively \cite{huang2024largelanguagemodelsselfcorrect}. Additionally, tests serve as (imperfect) verifiers that can later assist with selecting between candidate edits (see Section~\ref{sec:selection}).

Empirically, we find that models often require several iterations in order to write a functional test (e.g. because of configuration errors, easy-to-fix crashes, etc.). To allow models to focus on these steps first, we decompose this stage of our system into two back-to-back state machines: 

\begin{enumerate}
    \item An initial \textbf{testing state machine} which iterates on an initial draft of the test script.
    \item A follow-up \textbf{editing state machine} which iterates on a codebase edit (in the form of an aider-style edit diff \cite{aider}). This state machine is seeded with the output of a testing state machine and can also revise the testing script as needed.\footnote{Allowing models to continue revising tests during the editing state machine is important so that they can perform ``two-sided debugging''. In the testing state machine, models can keep iterating until their script correctly flags an issue when run on the unedited codebase. However, this can lead to tests that always report errors, even after a correct codebase edit has been applied. Allowing models to continue revising tests during the editing state machine lets them verify that their test fails pre-edit and passes post-edit.}
\end{enumerate}

The structure of these state machines is visualized in Figure~\ref{fig:sm_detail} (left and middle panels), with additional details given in Appendix~\ref{app:state_details}. Our decomposition into separate testing and editing state machines also lowers system costs. Our cost table (Table~\ref{tab:llm-costs}) shows that prefix cache reads are the most expensive component of the editing state machine, in large part due to the codebase files that are included in the initial prompt. Since writing a testing script generally does not require codebase context, we can reduce prompt lengths (and therefore cache read costs) in our testing state machine by omitting the files identified in Section~\ref{sec:context}. We additionally reduce prompt lengths by clearing the chat history between the testing and editing state machines.

We run 10 pairs of testing and editing state machines pairs for every instance in SWE-bench Verified and limit all state machines to eight iterations. On the left of Figure~\ref{fig:coverage_scaling}, we measure coverage as we sweep over both scaling parameters. Our best configuration uses all 10 state machines per instance and all eight iterations per state machine, achieving a coverage of \codemonkeysCoverage\%. Interestingly, after the first few iterations and state machines, a frontier emerges where configurations with similar total inference cost also have similar coverage, despite differences in how that cost is distributed across more state machines vs. more iterations. However, note that this does not mean that serial and parallel compute are fully interchangeable. In particular, two configurations with the same coverage will not necessarily obtain the same final score after selection. Scaling up parallel compute by generating many candidates can make selection more difficult, while generating a single, deep trajectory with a large number of iterations eliminates the selection problem entirely. Another distinction between the two types of scaling is that our setup only indirectly allows us to scale serial compute. We control the maximum number of iterations allowed within a single state machine. However, models can decide to approve of their test and/or edit and terminate a state machine before this iteration limit is reached.
Increasing the iteration limit further does not help ``stuck'' state machines where models have already incorrectly approved of their work. In contrast, we can always scale parallel compute further by running an additional state machine, guaranteeing a ``fresh start'' where a model may be able to generate a correct codebase edit.

\subsection{Selecting Between Candidate Edits}
\label{sec:selection}

\begin{figure*}[t]
    \begin{minipage}[t]{0.5\textwidth}
        \vspace{0pt}
        \includegraphics[width=\textwidth]{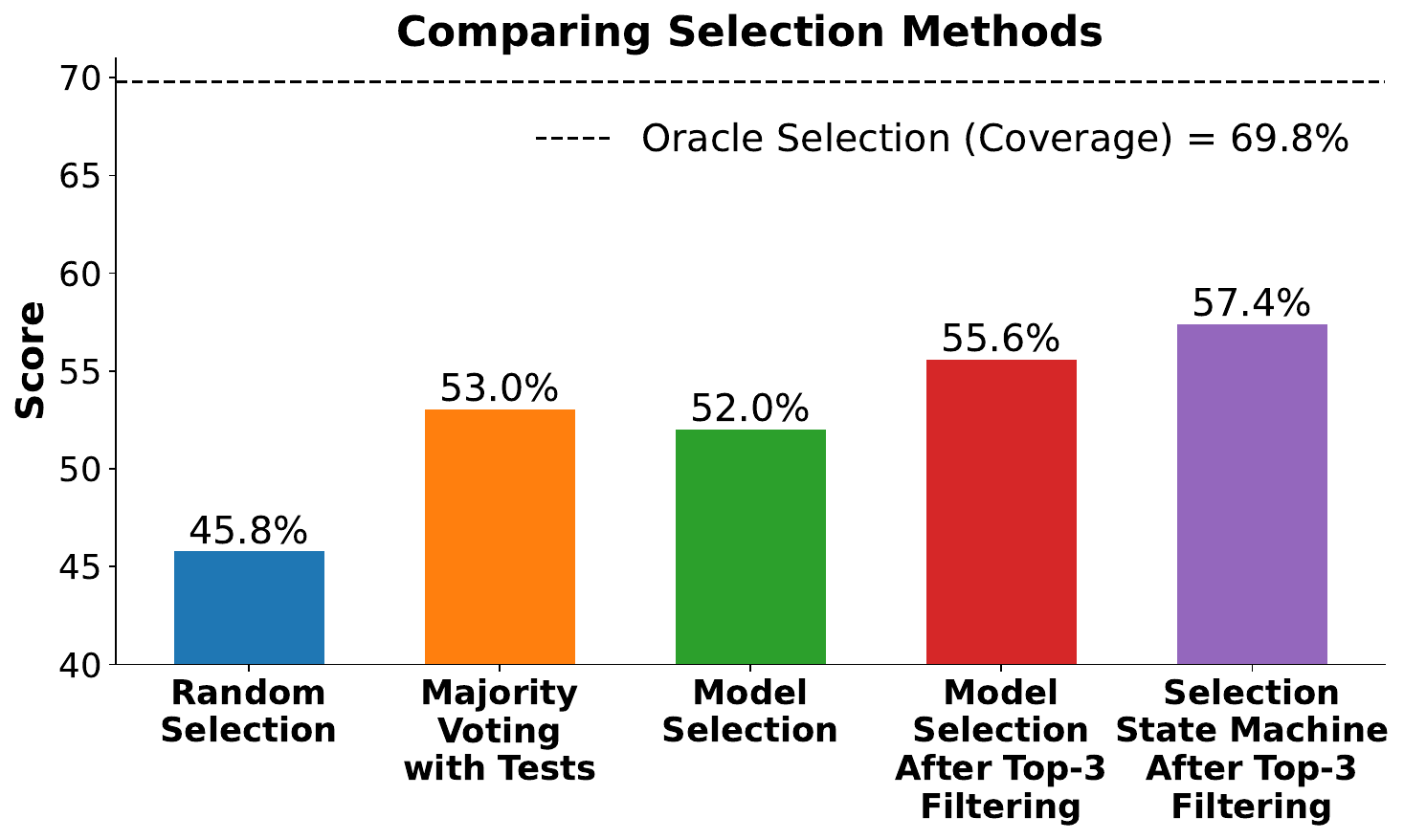}
        \caption{Comparing our selection methods when applied to the candidate edits generated by the CodeMonkeys editing state machines. Our best performing selection method -- the selection state machine after top-3 filtering with generated tests -- recovers approximately half of the difference between the random selection floor and the oracle selection ceiling (i.e. coverage).}
        
        \label{fig:selection_methods}
    \end{minipage}
    \hfill
    \begin{minipage}[t]{0.45\textwidth}
        \vspace{0pt}  
        \centering
        \small

\footnotesize
\begin{tabular}{@{}lcr@{}}
\toprule
\textbf{Method} & \textbf{Score} \\
\midrule
\textbf{Barrel of Monkeys  (Oracle Selection)} & 80.8 \\
o3 & 71.7 \\
\textbf{CodeMonkeys (Oracle Selection)} & 69.8 \\
\textbf{Barrel of Monkeys}  & 66.2 \\
Blackbox AI Agent & 62.8 \\
CodeStory & 62.2 \\
Learn-by-interact & 60.2 \\
devlo & 58.2 \\
\textbf{CodeMonkeys } & 57.4 \\
Emergent E1 & 57.2 \\
Gru & 57.0 \\

\bottomrule
\end{tabular}
\normalsize
\captionof{table}{Comparing final SWE-bench Verified scores between the methods explored in this paper (bolded) and existing top approaches. Note that the Barrel of Monkeys results rely on the generations from existing submissions on the SWE-bench Verified leaderboard, and oracle selection methods are coverage numbers.}
\label{tab:scores}
\end{minipage}
\vspace{-0.3cm}
\end{figure*}

Under a best-case scenario with oracle selection (where our final score equals our coverage of \codemonkeysCoverage\%), CodeMonkeys would outperform all submissions on the SWE-bench Verified leaderboard and trail o3’s reported score of \oThreeScore\% by only 1.9\%. If, however, we instead selected edits at random, our expected score of only \codemonkeysRandomScore\% would not even rank among the top 20 leaderboard submissions. This significant performance gap underscores the importance of accurate selection. We explore four strategies for selecting among the 10 candidate edits that we generate per instance:

\begin{enumerate}
    \item \textbf{Majority Voting with Tests:} We run each of the 10 model-generated tests on each of the 10 edits and select the edit that passes the most tests. If multiple edits tie with the most passes, we compute the expected score when picking randomly among them.
    
    \item \textbf{Model Selection:} We use a model to select a candidate edit after prompting it with the issue description, codebase context, and candidate edits in git diff form.
    
    \item \textbf{Model Selection After Top-3 Filtering:} We reuse the model selection prompt above, but select among only the three edits that pass the most generated tests. We break ties by favoring edits with shorter git diffs.   
    
    \item \textbf{Selection State Machine After Top-3 Filtering:} We upgrade the single-turn model-based selection approach to a full state machine that allows the model to write new testing scripts to differentiate between candidate edits (Figure~\ref{fig:sm_detail} on the right and Appendix~\ref{app:state_details}). The initial prompt for this state machine includes the same information as for model-based selection, but also includes an example model-generated test from the earlier stages. At each iteration, the model can either select a final edit or generate a new test. Whenever a new test is written, we show the model the outputs when running the test on every candidate edit in addition to the unedited codebase. We reuse the top-3 filtering process described above to narrow down the initial pool of candidates.
\end{enumerate}

We compare the performance of these selection methods in Figure~\ref{fig:selection_methods}. All four methods outperform random selection, with the selection state machine performing best. Note that the selection state machine is also the most expensive of the four methods we test; however, it contributes less than 10\% to total system costs. We also see the benefit of the initial filter with majority voting: model selection without top-3 filtering underperforms pure majority voting, while model selection with filtering outperforms it. With our chosen approach of selection state machine + top-3 filtering, CodeMonkeys achieves a final score of \codemonkeysScore\% on SWE-bench Verified (Table~\ref{tab:scores}), closing approximately half of the gap between random and oracle selection.

\subsubsection{Barrel of Monkeys: Selecting over an Ensemble of Samples from Existing SWE-bench Submissions}

In CodeMonkeys, we select among IID candidate edits generated by our method's state machines. Here, we demonstrate that our selection state machine is additionally helpful when combining candidate edits coming from heterogeneous sources. We create an ensemble of edits -- the ``Barrel of Monkeys'' -- by combining the final (already selected) edits from CodeMonkeys with the submissions from the top four entries on the SWE-bench Verified leaderboard\footnote{As of January 15, 2025.}: Blackbox AI Agent~\cite{blackbox}, CodeStory Midwit Agent + swe-search~\cite{codestory}, Learn-by-interact~\cite{su2025learnbyinteractdatacentricframeworkselfadaptive}, and devlo~\cite{devlo}. The ensemble, with five samples per problem, has a combined coverage of \barrelCoverage\%, which is notably higher than o3's reported score of \oThreeScore\%. While these two numbers are not directly comparable (o3's score corresponds to pass@1 while the Barrel of Monkeys' coverage corresponds to pass@5), we consider it important to highlight that (collectively) existing approaches can solve a significant fraction of instances from SWE-bench Verified. This further underscores the potential benefits of developing stronger methods for selection.

Since the Barrel of Monkeys begins with fewer initial candidates per problem than CodeMonkeys, we skip our initial test-based filtering and directly pass all edits to the selection state machine. The state machine's example test comes from the CodeMonkeys candidate edit that is part of the ensemble. Selecting over the ensemble achieves a score of \barrelScore\%, outperforming the best-performing member of the ensemble in isolation (Blackbox AI Agent \cite{blackbox} with a score of \bestScoreInEnsemble\%). However, since randomly selecting from this ensemble yields a score of \barrelRandomScore\%, our selection method recovers a smaller proportion of the gap between random selection and coverage here relative to selecting from the editing state machine.

\section{Limitations and Future Work}

In this work, we present the design of a system that successfully scales serial and parallel test-time compute in order to improve instance resolution rates on SWE-bench Verified. However, Figure~\ref{fig:problem_resolution_flow} demonstrates that there is still room to improve each stage of our system:

\begin{itemize}
    \item \textbf{Context:} Our file-level filter-and-rank procedure still omits relevant files for 7.4\% of SWE-bench Verified instances. As models’ usable context lengths grow and long-context processing becomes more efficient, we are hopeful that eventually this entire stage of our pipeline can be replaced with simply providing the entire codebase as context along with all relevant documentation \cite{magic2024context}.
    Another factor to consider is that the base models already possess background knowledge on the repositories they are editing.
    This assumption allows us to avoid needing to provide the model with a more fundamental explanation about the purpose and usage of the package being edited (e.g. through documentation).
    This assumption will not hold when solving issues from new or private repositories that are not well-represented in the model training data, which will likely require modifications to our retrieval pipeline to attain high recalls.

    \item \textbf{Generating Edits and Tests:} We also see room for improving the CodeMonkeys state machines and further increasing coverage. In our state machines, we only provide models with execution feedback from their testing scripts, with each script attempting to reproduce the target issue. Additional execution feedback could come from model-written or existing regression tests that help prevent new bugs from being introduced. Moreover, like in Large Language Monkeys~\cite{monkeys}, CodeMonkeys incorporates diversity into generation exclusively through using a positive token sampling temperature. Distinct pairs of testing/editing state machines have no awareness of other state machines that are being made for the same problem, which can lead to redundant/non-diverse generations. Alternative approaches that add serial dependencies and inform models of previous attempts~\cite{plansearch} could encourage greater diversity during generation.

    \item \textbf{Selection:} Our approach to selection recovers roughly half of the score gap between random and oracle selection when applied to CodeMonkeys, and an even smaller fraction of the gap when selecting over the Barrel of Monkeys. Improvements to selection methods, even without any changes to the candidate generation procedure, can therefore significantly raise overall SWE-bench scores. Similar to the potential improvements listed for generation, one source of selection signal that we do not exploit is the existing suite of tests inside of each repo. Agentless \cite{xia2024agentlessdemystifyingllmbasedsoftware} and the Gemini team \cite{flash2} both include these tests in their selection procedures.
\end{itemize}

Moreover, our work omits additional approaches to scaling test-time compute, such as ensembling different models together \cite{saadfalcon2024archonarchitecturesearchframework, wang2024mixtureofagentsenhanceslargelanguage} and leveraging dedicated reasoning models which are trained to explore the space of candidate solutions in an extended chain of thought \cite{o1, qwq}. 
Overall, we are excited about continued progress in methods for scaling test-time compute and systems that harness this scaling to solve real-world tasks.

\section{Related Work}

\textbf{AI for Software Engineering:} There has been considerable interest in applying LLMs to coding tasks \cite{chen2021evaluatinglargelanguagemodels, li2023starcodersourceyou, liu2024largelanguagemodelbasedagents}. Progress in this area can be measured with a diverse set of benchmarks that test models’ abilities to complete functions from a prompt \cite{chen2021evaluatinglargelanguagemodels, austin2021programsynthesislargelanguage}, build entire libraries from a specification \cite{zhao2024commit0librarygenerationscratch}, and perform domain-specific programming tasks \cite{ouyang2024kernelbench, tian2024scicoderesearchcodingbenchmark}.
In our work, we focus on SWE-bench~\cite{jimenez2024swebenchlanguagemodelsresolve}, a dataset of real-world GitHub issues from popular Python repositories. The
state-of-the-art on this benchmark has rapidly improved since its release: initial methods resolved less than 5\% of instances from SWE-bench Verified \cite{chowdhury2024swebenchverified}, while current approaches can solve more than 60\% \cite{blackbox, codestory, su2025learnbyinteractdatacentricframeworkselfadaptive}.
This improvement can be attributed to stronger underlying models \cite{claude35sonnet, o1, flash2} in addition to better frameworks for equipping models with tools and guiding the issue resolution process.

These frameworks occupy a large design space. Some methods adopt a relatively hands-off approach that provide models with a suite of tools that they can use in an arbitrary order until they have resolved the issue \cite{yang2024sweagentagentcomputerinterfacesenable, schluntz2024raising}. Other approaches (including CodeMonkeys) enforce a more strict structure on the issue resolution process, e.g. by introducing state machines \cite{moatless} or a dedicated sequence of steps \cite{xia2024agentlessdemystifyingllmbasedsoftware}. Similar to CodeMonkeys, Agentless \cite{xia2024agentlessdemystifyingllmbasedsoftware} partitions issue resolution into steps for identifying relevant context, generating candidate edits, and selecting between these edits. However, our implementation of each of these steps (e.g. using a LLM-powered scan to identify context and using multi-turn feedback loops for generating edits and selecting between them) is distinct from Agentless.

Frameworks also differ in the types of tools that they provide to models. Some tools are general purpose, such as the ability to run shell commands or search the web \cite{yang2024sweagentagentcomputerinterfacesenable, schluntz2024raising, wang2024openhandsopenplatformai}. Other tools are more narrow, such as editing files with the search-and-replace format introduced by Aider \cite{aider, moatless, xia2024agentlessdemystifyingllmbasedsoftware}.
Some frameworks also provide models with tools for identifying relevant codebase context, such as by opening files or running a semantic search \cite{moatless, yang2024sweagentagentcomputerinterfacesenable, schluntz2024raising}.
Existing frameworks for solving SWE-bench issues have also explored scaling test-time compute. Serial compute is commonly scaled by asking models to reflect/revise their work \cite{moatless} or by providing them with execution or tool call feedback \cite{wang2024openhandsopenplatformai, huang2024agentcodermultiagentbasedcodegeneration, ruan2024specrovercodeintentextraction}. Notably, with Anthropic's framework for showcasing the updated Claude 3.5 Sonnet \cite{schluntz2024raising}, Claude sometimes used hundreds of turns of feedback before submitting a correct solution. 
Agentless \cite{xia2024agentlessdemystifyingllmbasedsoftware} and the Gemini team \cite{flash2} scale parallel compute by generating multiple candidate edits and using repositories’ existing unit tests to help select between them. Agentless additionally uses model-written reproduction tests during selection to check that candidate edits actually resolved the issue, while the Gemini team and SpecRover \cite{ruan2024specrovercodeintentextraction} incorporate model-based selection. 
Methods like SWE-search~\cite{antoniades2024swesearchenhancingsoftwareagents} and CodeStory~\cite{codestory} combine serial and parallel scaling by performing a tree search over intermediate states using model-based value estimation.

\textbf{Scaling Test-Time Compute:} Expending test-time compute via tree search has long been a successful strategy when designing AI systems to play games~\cite{deepblue, silver2017mastering, pluribus}. Recently, similar search techniques have also been combined with LLMs to successfully prove formal mathematical statements \cite{Trinh2024alphageometry}. Across a wider variety of settings, allowing LLMs to use more tokens to think through a problem before outputting a final answer has led to large boosts in model reasoning capabilities~\cite{wei2023chainofthoughtpromptingelicitsreasoning, nye2021workscratchpadsintermediatecomputation}. Explicitly optimizing models to perform this deliberation process (e.g. with reinforcement learning) has improved capabilities even further~\cite{o1, qwq, deepseekai2025deepseekr1incentivizingreasoningcapability}.

Existing work has also explored scaling parallel test-time compute by sampling multiple completions per problem~\cite{monkeys, wu2024inferencescalinglawsempirical, snell2024scalingllmtesttimecompute}. 
Across math and coding tasks, repeatedly sampling with small models can obtain higher coverage than single samples from a larger model \cite{hassid2024largerbetterimprovedllm}. Repeated sampling can be particularly effective with coding tasks since the ability to run and test model outputs can assist with selection \cite{Li_2022_alphacode, arc_gpt4o, li2024combininginductiontransductionabstract}. More general approaches to selection include using outcome reward models \cite{christiano2017deepreinforcementlearninghuman}, process reward models \cite{lightman2023letsverifystepstep, wang2024mathshepherdverifyreinforcellms}, or prompting-based verifier setups \cite{snell2024scalingllmtesttimecompute}. PlanSearch~\cite{plansearch} shows that converting some parallel samples into sequential attempts can improve diversity, reducing the compute needed to achieve a given coverage.
Archon~\cite{saadfalcon2024archonarchitecturesearchframework}, Mixture-of-Agents~\cite{wang2024mixtureofagentsenhanceslargelanguage} and MALT \cite{motwani2024malt} consider the setting where many different models can generate samples or act as sample verifiers and aggregators, exploring how performant combinations of these LLM components can be automatically discovered. 

\section{Acknowledgements}

We are grateful to Benjamin Spector, Chris Fifty, Jerry Liu, Jon Saad-Falcon, Owen Dugan, Quinn McIntyre, Simon Guo, and Will Tennien for their helpful discussions and feedback throughout this project.
% We also thank Bailey, Marley, Charlie and Poppy for being good dogs.

We gratefully acknowledge the support of NIH under No. U54EB020405 (Mobilize), NSF under Nos. CCF2247015 (Hardware-Aware), CCF1763315 (Beyond Sparsity), CCF1563078 (Volume to Velocity), and 1937301 (RTML); US DEVCOM ARL under Nos. W911NF-23-2-0184 (Long-context) and W911NF-21-2-0251 (Interactive Human-AI Teaming); ONR under Nos. N000142312633 (Deep Signal Processing); Stanford HAI under No. 247183; NXP, Xilinx, LETI-CEA, Intel, IBM, Microsoft, NEC, Toshiba, TSMC, ARM, Hitachi, BASF, Accenture, Ericsson, Qualcomm, Analog Devices, Google Cloud, Salesforce, Total, the HAI-GCP Cloud Credits for Research program,  the Stanford Data Science Initiative (SDSI), and members of the Stanford DAWN project: Meta, Google, and VMWare. The U.S. Government is authorized to reproduce and distribute reprints for Governmental purposes notwithstanding any copyright notation thereon. Any opinions, findings, and conclusions or recommendations expressed in this material are those of the authors and do not necessarily reflect the views, policies, or endorsements, either expressed or implied, of NIH, ONR, or the U.S. Government. 

This work was completed with the support of the Clarendon Fund Scholarships.

\bibliographystyle{plain}
\bibliography{main}

\appendix
\newpage

\section{DeepSeek-V3 Results}
\label{app:deepseek}

\begin{figure*}[t]
    \centering
    \includegraphics[width=\textwidth]{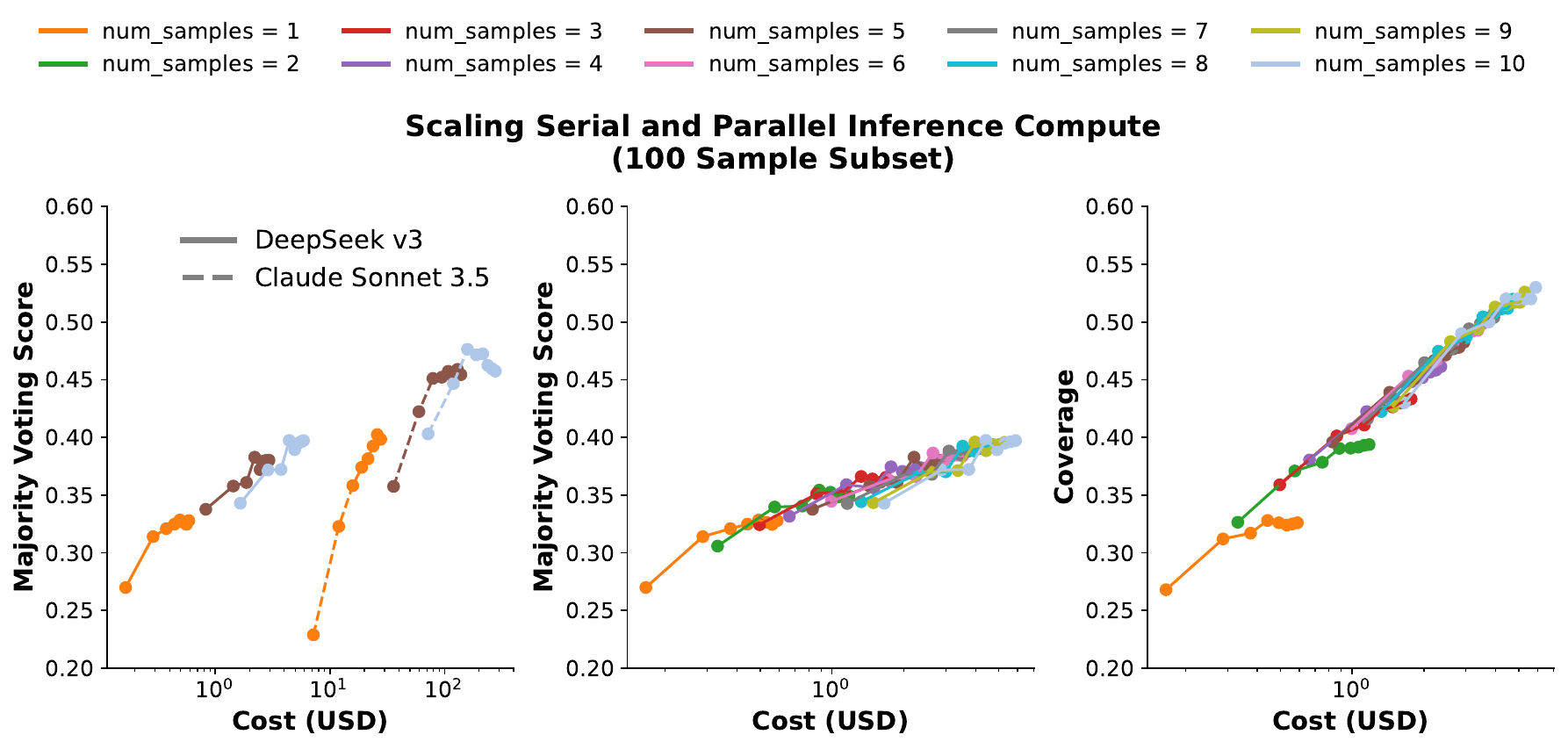}
    \caption{\textbf{Left:} Comparing the impact of scaling the number of parallel samples and sequential iterations on majority voting score between Claude and DeepSeek-V3. Each line corresponds to a fixed amount of samples, with each dot on the line being a different maximum number of sequential iterations. We see that although Claude can achieve a higher overall score, DeepSeek-V3 can achieve $86.8\%$ percent of the score at a fraction of the cost. \textbf{Center:} A more granular view of the majority voting scaling for DeepSeek-V3. \textbf{Right:} Coverage for DeepSeek-V3 as a function of the number of parallel samples and sequential iterations. We highlight that coverage is continuing to scale with increased inference compute.}
    \label{fig:deepseek_maj_voting_score_scaling}
    \vspace{-0.5cm}
\end{figure*}

Here, we conduct an initial evaluation of DeepSeek-V3 \cite{deepseekai2024deepseekv3technicalreport} as a potential substitute for our system’s use of Claude 3.5 Sonnet. We reuse the same codebase context files as our primary CodeMonkeys experiments and rerun our testing and editing state machines on a 100-instance random subset of SWE-bench Verified using DeepSeek-v3. 
In the Figure~\ref{fig:deepseek_maj_voting_score_scaling}, we compare how coverage and majority voting score scale with the number of samples and iterations when compared to Claude Sonnet 3.5.
We note that Claude Sonnet 3.5 is able to achieve a score of $45.74\%$ on this subset of problems, which is $6.02\%$ higher than the best score of DeepSeek-V3.
However, the DeepSeek API is over an order of magnitude cheaper than that of Claude.
These results highlight the potential benefits of generating many candidate solutions from a cheaper model like DeepSeek-V3 so long as a selection method can identify correct samples from large collections.

\section{Experimental Details}

We release our code and trajectories at \url{https://scalingintelligence.stanford.edu/pubs/codemonkeys/}. This release includes:
\begin{itemize}
    \item All the code needed to run CodeMonkeys on SWE-bench Verified.
    \item All the commands needed to generate the figures/tables in this paper.
    \item The complete trajectories taken by CodeMonkeys when solving each problem.
\end{itemize}

\subsection{State Machine Details}
\label{app:state_details}

All three CodeMonkeys state machines (the testing, editing, and selection state machines) follow the same structure.
Each state machine begins by giving the model an initial prompt and an initial task. Once the model completes this initial task, a feedback loop is entered. At each iteration, the model is provided with some execution feedback from the previous iteration. The model is then allowed to either revise its work, triggering a new iteration of the loop, or approve its work, terminating the state machine.
In Table~\ref{tab:state-machines}, we provide the inputs, initial task, information given at each iteration, and the task at each iteration for all three state machines.
For full prompts, please see the folder \texttt{codemonkeys/prompts} in our codebase.

\begin{table}[h]
\centering
\small
\begin{tabular}{L{2cm}L{4.2cm}L{4.2cm}L{4.2cm}}
\toprule
& \thead{Testing\\State Machine} & \thead{Editing\\State Machine} & \thead{Selection\\State Machine} \\
\midrule
\textbf{Information in Initial Prompt} & 
GitHub issue description. & 
GitHub issue description, codebase context, final test script from a testing state machine, execution output when running the provided test on the unedited codebase. & 
GitHub issue description, candidate edits in git diff form, full contents of any codebase files that have been edited, test script from the edit that passed the most generated tests (breaking ties by using the shortest edit). \\
\midrule
\textbf{Initial Task} & 
Write a test script that reproduces the issue. The script should exit with code 0 if the issue is fixed and exit with code 2 if the issue is not fixed. & 
Write a codebase edit that resolves the issue. & 
Write a test script for distinguishing between candidates and assessing their correctness. \\
\midrule
\textbf{Information in Iteration Prompt} & 
Execution output when running the test on the codebase (which has not yet been edited). & 
Execution outputs when running the testing script on the unedited codebase and edited codebase. & 
Execution outputs when running the test script on a codebase after each edit has been applied, in addition to the unedited codebase. \\
\midrule
\textbf{Iteration Task} & 
Rewrite the test script or approve of it (terminating the state machine). & 
Rewrite the edit, rewrite the test script, or approve of the (edit, test) pairs (terminating the state machine). & 
Write a new testing script, or make a selection among the candidate edits (terminating the state machine). \\
\bottomrule
\end{tabular}
\caption{Details of the Testing, Editing, and Selection State Machines.}
\label{tab:state-machines}
\end{table}

\subsection{Hyperparameters}

\noindent{We elaborate on other experimental details for each part in Table~\ref{tab:hyperparams}}.

\begin{table}[htbp] \centering   

\begin{tabular}{llcr}
\toprule
\textbf{Subtask} & \textbf{Stage} & \textbf{Parameter} & \textbf{Value} \\
\midrule
Context & Relevance & Model & Qwen-2.5-Coder-32B-Instruct \\
& & Hardware & 8xL40S \\
& & Temperature & 0.0 \\
\cmidrule(lr){2-4}
& Ranking & Model & Claude Sonnet 3.5 \\
& & Temperature & 0.0 \\
& & Repetitions & 3 \\
\midrule
Generation & Testing \& Editing & Temperature (Sonnet 3.5) & 0.5 \\
& & Temperature (DeepSeek-V3) & 0.6 \\
& & Number of State Machines per Instance & 10 \\
& & Maximum Iterations & 8 \\
& & Generated Test Timeout (seconds) & 100 \\
\midrule
Selection & --- & Temperature & 0.0 \\
& & Maximum Iterations & 10 \\
& & Generated Test Timeout (seconds) & 100 \\
\bottomrule
\end{tabular}

\caption{CodeMonkeys hyperparameter summary.}
\label{tab:hyperparams} \end{table}

\section{Local Compute for Relevance}
\label{sec:local_costs}
For the relevance stage, we run Qwen-2.5-Coder-32B-Instruct \cite{hui2024qwen25coder} locally in bf16 across 8xL40S GPUs. Each L40S has a bf16 compute throughput of $362.05$ TFLOPS \cite{nvidia2024l40s}, giving our node a theoretical throughput of:

\begin{equation*}
    8 \text{ devices} \cdot 362.05 \text{ TFLOPS/device} = 2{,}896.4 \text{ TFLOPS}
\end{equation*}

\noindent{Since the model we run is a dense transformer with no parameter sharing, we can estimate its FLOPs per token as 2 * num\_parameters \cite{sardana2024chinchillaoptimalaccountinginferencelanguage}. Assuming a hardware utilization of 20\% during inference, we obtain a per-node throughput of}:

$$\text{Throughput} \approx \frac{0.2 \cdot 2{,}896.4 \cdot 10^{12} \text{ FLOPs/second}}{2 \cdot 32 \cdot 10^9 \text{ FLOPs/token}} = 9{,}051 \text{ tokens/second}$$

\noindent{Across the 500 instances in SWE-bench Verified, the relevance stage requires processing a total of $1.32084 \cdot 10^9$ tokens. To estimate the cost of compute,  we use RunPod's pricing: an 8xL40S node costs \$8.24 per hour \cite{runpod2024}, yielding a total cost of:}

\begin{align*}
    \text{Relevance cost} &\approx \frac{1.32083 \cdot 10^9 \text{ tokens}}{9{,}051 \text{ tokens/second} \cdot 3{,}600 \text{ seconds/hour}} \cdot \$8.24/\text{hour} \\
    &= 40.5 \text{ hours} \cdot \$8.24/\text{hours} \\
    &= \$334.02
\end{align*}

\end{document}